\documentclass[conference]{IEEEtran}
\IEEEoverridecommandlockouts

\usepackage{cite}
\usepackage{amsmath,amssymb,amsfonts}

\usepackage{algorithm}
\usepackage{algpseudocode}
\usepackage{graphicx}
\usepackage{textcomp}
\usepackage{xcolor}
\usepackage{booktabs}
\usepackage{multirow}
\usepackage{array}
\usepackage{colortbl}
\usepackage{hyperref}
\usepackage{scalerel}     
\usepackage{orcidlink}    

\def\BibTeX{{\rm B\kern-.05em{\sc i\kern-.025em b}\kern-.08em
    T\kern-.1667em\lower.7ex\hbox{E}\kern-.125emX}}

\begin{document}

\title{Semi-Supervised Supply Chain Fraud Detection with Unsupervised Pre-Filtering}

\author{
\IEEEauthorblockN{Fatemeh Moradi\,\orcidlink{0009-0001-6077-7503}}
\IEEEauthorblockA{Faculty of Engineering \\
Isfahan (Khorasgan) Branch,\\ Islamic Azad University \\
Isfahan, Iran}
\and
\IEEEauthorblockN{Mehran Tarif\,\orcidlink{0009-0003-0951-1800}}
\IEEEauthorblockA{Department of Computer Science \\
University of Verona \\
Verona, Italy}
\and
\IEEEauthorblockN{Mohammadhossein Homaei\,\orcidlink{0000-0002-6108-6632}}
\IEEEauthorblockA{Media Engineering Group \\
University of Extremadura \\
Cáceres, Spain}
}

\maketitle

\begin{abstract}
Detecting fraud in modern supply chains is a growing challenge, driven by the complexity of global networks and the scarcity of labeled data. Traditional detection methods often struggle with class imbalance and limited supervision, reducing their effectiveness in real-world applications. This paper proposes a novel two-phase learning framework to address these challenges. In the first phase, the Isolation Forest algorithm performs unsupervised anomaly detection to identify potential fraud cases and reduce the volume of data requiring further analysis. In the second phase, a self-training Support Vector Machine (SVM) refines the predictions using both labeled and high-confidence pseudo-labeled samples, enabling robust semi-supervised learning. The proposed method is evaluated on the DataCo Smart Supply Chain Dataset, a comprehensive real-world supply chain dataset with fraud indicators. It achieves an F1-score of 0.817 while maintaining a false positive rate below 3.0\%. These results demonstrate the effectiveness and efficiency of combining unsupervised pre-filtering with semi-supervised refinement for supply chain fraud detection under real-world constraints, though we acknowledge limitations regarding concept drift and the need for comparison with deep learning approaches.

\end{abstract}

\section{Introduction}
Supply chain fraud has become a critical threat to global commerce, with organizations facing increasingly sophisticated fraudulent schemes across complex supply networks \cite{modruvsan2021review}. The digitalization of supply chains has created new vulnerabilities, enabling various fraudulent activities including procurement fraud, vendor impersonation, invoice manipulation, and counterfeit goods infiltration \cite{constante2019fraud}. These activities result in substantial financial losses and operational disruptions for businesses worldwide \cite{baryannis2019predicting}. The interconnected nature of modern supply chains amplifies fraud impact, as single fraudulent events can cascade across multiple organizations and geographical regions, while increasing transaction complexity challenges traditional real-time monitoring systems.
Modern supply chains involve multiple stakeholders, heterogeneous data sources, and intricate networks that complicate fraud detection \cite{zhou2021supply}. The inherent class imbalance and sophisticated fraud schemes have rendered traditional rule-based systems inadequate \cite{bauder2018effects,phua2010comprehensive}.
Zhou et al. \cite{zhou2021supply} demonstrated XGBoost's effectiveness for supply chain fraud prediction, emphasizing feature engineering and ensemble methods. The DataCo Supply Chain Dataset \cite{dataco2019} has become a valuable benchmark for fraud detection research. Constante-Nicolalde et al. \cite{constante2019fraud} explored smart supply chain fraud prediction with IoT integration. Baryannis et al. \cite{baryannis2019predicting} examined the performance-interpretability trade-off in machine learning for supply chain risk prediction.

To address these limitations, innovative fraud detection systems must effectively handle limited labeled data while maintaining high detection accuracy and computational efficiency through hybrid approaches combining multiple learning paradigms. Prior work has explored artificial intelligence's role in enhancing cybersecurity across digital infrastructures, including digital twin systems, emphasizing hybrid AI techniques for detecting complex and evolving threats \cite{Homaei2022RECSI,Homaei2024,Moradi2025review}.
This paper proposes a novel two-phase learning model combining unsupervised anomaly detection with semi-supervised learning refinement. The first phase employs Isolation Forest for efficient outlier identification without requiring labeled training data \cite{liu2008isolation}. The second phase utilizes self-training Support Vector Machine to refine detection results by iteratively expanding the labeled dataset with high-confidence predictions \cite{amini2025self}. This approach addresses key challenges including computational efficiency, class imbalance handling, and effective utilization of limited labeled data in supply chain fraud detection systems.
The rest of this paper is organized as follows. Section~\ref{sec2} reviews recent related works on fraud detection in supply chains and machine learning methods. Section~\ref{sec3} explains the proposed two-phase model, including Isolation Forest and self-training SVM. Section~\ref{sec4} describes the experimental setup with datasets, evaluation metrics, and baseline methods. Section~\ref{sec5} presents and discusses the results. Finally, Section~\ref{sec6} gives the conclusion and suggestions for future work.

\section{Related Work}\label{sec2}
\subsection{Supply Chain Fraud Detection}
Supply chain fraud detection has gained significant attention due to increasing complexity and digitalization of global networks. Modruš­an et al. \cite{modruvsan2021review} reviewed public procurement fraud detection techniques, highlighting evolution from rule-based to machine learning approaches and identifying key challenges: data heterogeneity, real-time processing, and sophisticated fraud schemes.
Zhou et al. \cite{zhou2021supply} demonstrated XGBoost's effectiveness for supply chain fraud prediction, while the DataCo dataset \cite{dataco2019} has become a valuable benchmark. Recent work has explored IoT integration \cite{constante2019fraud} and performance-interpretability trade-offs \cite{baryannis2019predicting}.

\subsection{Machine Learning Approaches for Fraud Detection}
Machine learning applications in fraud detection have been extensively surveyed. Hernández Aros et al. \cite{hernandez2024financial} reviewed financial fraud detection literature, analyzing 104 articles and identifying Random Forest and Autoencoder as particularly effective techniques. Phua et al. \cite{phua2010comprehensive} provided a comprehensive survey of data mining-based fraud detection across multiple domains. Recent work has applied semi-supervised learning using Isolation Forests to effectively detect fraud in supply chain data without full supervision \cite{liu2023semi}.
\subsection{Class Imbalance in Fraud Detection}
Class imbalance represents a significant challenge in fraud detection. Bauder and Khoshgoftaar \cite{bauder2018effects} investigated varying class distribution effects on learner behavior for Medicare fraud detection, demonstrating that unsupervised learning approaches can offer advantages with severely imbalanced datasets \cite{Moradi2025ensemble}. Wei et al. \cite{wei2013effective} addressed sophisticated online banking fraud detection on extremely imbalanced data (<0.1\% fraud rate), combining multiple techniques to handle extreme imbalance while maintaining high detection accuracy.

\subsection{Isolation Forest and Anomaly Detection}
Isolation Forest \cite{liu2008isolation} represents a paradigm shift in anomaly detection, using isolation principles rather than distance or density-based measures. The algorithm's insight that anomalies are "few and different" enables efficient detection with O(n log n) complexity, suitable for large-scale supply chain fraud detection. Liu et al. \cite{liu2012isolation} provided detailed analysis of the algorithm's performance characteristics and robustness properties. Hariri et al. \cite{hariri2019extended} proposed Extended Isolation Forest, addressing bias issues by using hyperplanes with random slopes, improving detection consistency and accuracy.
\subsection{Semi-supervised Learning and Self-training}
Semi-supervised learning approaches show promise for fraud detection with scarce labeled data. Wang et al. \cite{wang2019semi} developed a semi-supervised graph attentive network for financial fraud detection with substantial improvements. Hyun et al. \cite{hyun2020class} proposed Suppressed Consistency Loss (SCL) to handle distribution differences between labeled and unlabeled data. Wei et al. \cite{wei2021crest} introduced CReST for imbalanced semi-supervised learning, achieving 11.8\% improvement over FixMatch. Amini et al. \cite{amini2025self} surveyed self-training methodologies, providing guidance for selecting appropriate strategies. One-Class SVM has been employed to model normal transaction behavior in supply chains \cite{wang2022fraud}.
\subsection{Research Gap}
Despite extensive research in fraud detection, existing approaches face three critical limitations in supply chain contexts: (1) supervised methods require extensive labeled data that is costly to obtain, (2) unsupervised methods suffer from high false positive rates when used in isolation, and (3) current semi-supervised approaches do not address the computational scalability required for real-time supply chain monitoring. Our work addresses this gap by proposing a computationally efficient two-phase framework that combines the strengths of unsupervised and semi-supervised learning while maintaining practical deployment feasibility.

\section{Methodology}\label{sec3}

\subsection{Problem Formulation}

Let $\mathcal{X} = \{x_i\}_{i=1}^n$ represent the complete supply chain transaction dataset (such as the DataCo dataset \cite{dataco2019}), where $x_i \in \mathbb{R}^d$ denotes the $d$-dimensional feature vector for transaction $i$, and $n$ is the total number of transactions. Each transaction has an associated true label $y_i \in \{0, 1\}$ (0 for legitimate, 1 for fraudulent), but these labels are only observed for a small subset of the data.

The dataset can be partitioned into two disjoint subsets based on label availability:
\begin{itemize}
    \item Labeled subset: $\mathcal{D}_L = \{(x_i, y_i)\}_{i=1}^{n_L}$, where $n_L \ll n$
    \item Unlabeled subset: $\mathcal{D}_U = \{x_i\}_{i=n_L+1}^{n}$, where labels exist but are unobserved
\end{itemize}

We denote the complete dataset as $\mathcal{D} = \mathcal{D}_L \cup \mathcal{D}_U$, where $|\mathcal{D}_L| = n_L$ and $|\mathcal{D}_U| = n - n_L$. In practical supply chain scenarios, the dataset exhibits severe class imbalance with $|\{i : y_i = 1, (x_i, y_i) \in \mathcal{D}_L\}| \ll |\{i : y_i = 0, (x_i, y_i) \in \mathcal{D}_L\}|$, where fraudulent transactions constitute a small minority of the labeled data.

The objective is to learn a classifier $f: \mathbb{R}^d \rightarrow \{0, 1\}$ that effectively identifies fraudulent transactions across the entire dataset $\mathcal{X}$, leveraging both the limited labeled data in $\mathcal{D}_L$ and the abundant unlabeled data in $\mathcal{D}_U$, while minimizing false positives and maintaining computational efficiency for real-time processing requirements.

\begin{algorithm}[htb]
\scriptsize
\caption{Two-Phase Learning Framework}
\label{alg:two_phase_framework}
\begin{algorithmic}[1]
\Require Dataset $\mathcal{D}$, labeled subset $\mathcal{D}_L$, unlabeled subset $\mathcal{D}_U$, 
         parameters $\alpha$, $\theta_{base}$, $\beta$
\Ensure Refined fraud detection model $f_{final}$

\State \textbf{Phase 1: Isolation Forest Pre-filtering}
\State Train Isolation Forest model $IF$ on entire dataset $\mathcal{D}$
\State Compute anomaly scores: $s_i = IF(x_i)$ for all $x_i \in \mathcal{D}$
\State Calculate threshold: $\tau = \mu_s + \alpha \sigma_s$
\State Create candidate set: $\mathcal{D}_{candidates} = \{x_i \in \mathcal{D}_U : s_i \geq \tau\}$

\State \textbf{Phase 2: Self-training SVM Refinement}
\State Initialize SVM classifier $SVM_0$ using labeled data $\mathcal{D}_L$
\State $\mathcal{D}_L^{(0)} \leftarrow \mathcal{D}_L$, $t \leftarrow 0$
\While{$\Delta F1_t \geq 0.001$ and $t < 10$}
    \State Predict on candidates: $\hat{y}_i = SVM_t(x_i)$ for $x_i \in \mathcal{D}_{candidates}$
    \State Compute confidence: $c_i = |f(x_i)|/\max_j|f(x_j)|$
    \State Calculate class-specific thresholds using Eq.~\ref{eq:class_balanced_threshold}
    \State Select high-confidence: $\mathcal{P}_t = \{(x_i, \hat{y}_i) : c_i \geq \theta_{\hat{y}_i}\}$
    \State Update labeled set: $\mathcal{D}_L^{(t+1)} = \mathcal{D}_L^{(t)} \cup \mathcal{P}_t$
    \State Remove from candidates: $\mathcal{D}_{candidates} \leftarrow \mathcal{D}_{candidates} \setminus \mathcal{P}_t$
    \State Retrain: $SVM_{t+1}$ on $\mathcal{D}_L^{(t+1)}$ with class weights
    \State $t \leftarrow t + 1$
\EndWhile

\State \Return $f_{final} = SVM_t$
\end{algorithmic}
\end{algorithm}

The algorithm terminates when either the F1-score improvement between iterations ($\Delta F1_t$) falls below 0.001 or the maximum number of iterations (10) is reached, ensuring convergence while preventing overfitting.

\subsection{Two-Phase Framework Overview}

Our proposed methodology consists of two sequential phases designed to address the key challenges in supply chain fraud detection: computational scalability, class imbalance, and limited labeled data availability. The framework architecture is illustrated in Algorithm \ref{alg:two_phase_framework}.
\begin{figure*}[t]
\centering
\includegraphics[width=0.7\textwidth]{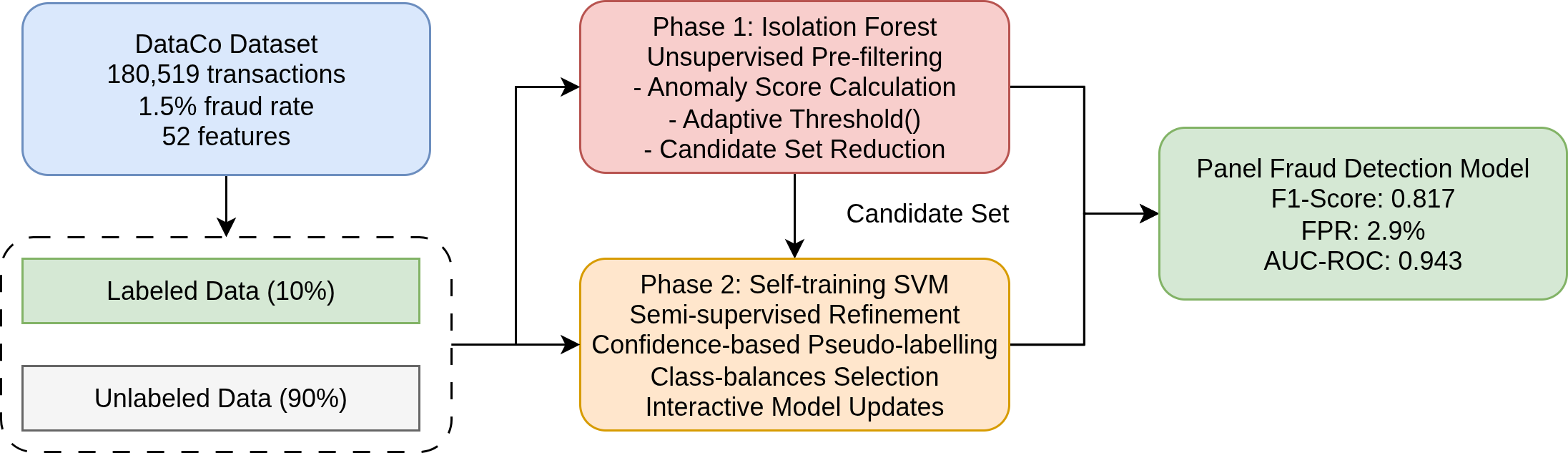}
\caption{Two-Phase Learning Framework combining Isolation Forest pre-filtering with self-training SVM refinement, achieving F1-score of 0.817 and 2.9\% false positive rate on DataCo dataset.}
\label{fig:framework}
\end{figure*}

The complete two-phase learning framework is illustrated in Figure 1, which provides a comprehensive overview of our proposed methodology. The framework begins with the DataCo dataset containing 180,519 transactions with a 1.5\% fraud rate. The data is partitioned into labeled (10\%) and unlabeled (90\%) subsets to simulate realistic semi-supervised scenarios. Phase 1 employs Isolation Forest for unsupervised pre-filtering to identify potential fraud candidates, while Phase 2 utilizes self-training SVM for semi-supervised refinement through confidence-based pseudo-labeling and iterative model updates (Figure~\ref{fig:framework}).

\subsection{Phase 1: Isolation Forest Pre-filtering}

The Isolation Forest algorithm, proposed by Liu et al. \cite{liu2008isolation}, operates on the principle that anomalies are easier to isolate than normal instances. For a given transaction $x_i$, the anomaly score is computed as:

\begin{equation}
s(x_i) = 2^{-\frac{E(h(x_i))}{c(n)}}
\label{eq:anomaly_score}
\end{equation}

where $E(h(x_i))$ represents the average path length of $x_i$ over all isolation trees, and $c(n)$ is the average path length of unsuccessful search in a Binary Search Tree (BST) with $n$ points:

\begin{equation}
c(n) = 2H(n-1) - \frac{2(n-1)}{n}
\label{eq:average_path}
\end{equation}

The threshold selection strategy employs an adaptive threshold approach:

\begin{equation}
\tau = \mu_s + \alpha \sigma_s
\label{eq:threshold}
\end{equation}

where $\mu_s$ and $\sigma_s$ represent the mean and standard deviation of anomaly scores, respectively, and $\alpha$ is a sensitivity parameter.

\begin{figure}[htb]
\centering
\includegraphics[width=\columnwidth]{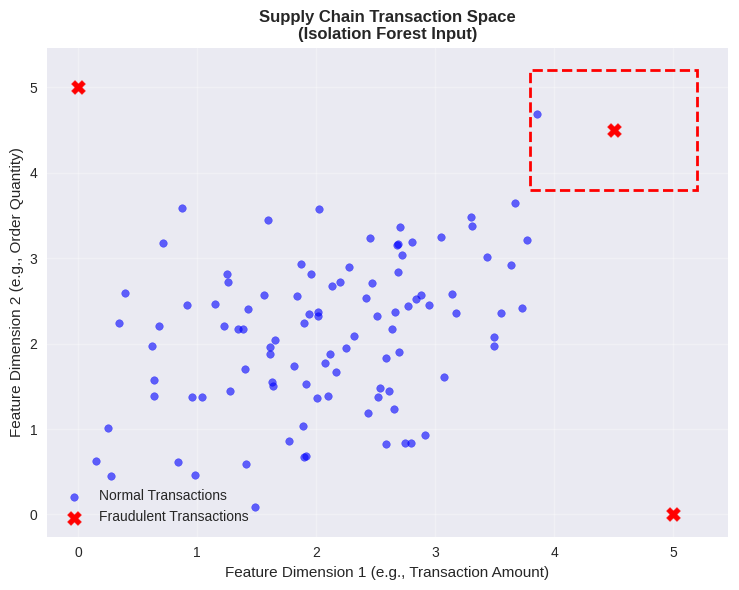}
\caption{As an example, the Supply Chain Transaction Space showing normal transactions (blue circles) clustering densely while fraudulent transactions (red X) appear as isolated outliers.}
\label{fig:transaction_space}
\end{figure}

Figure~\ref{fig:transaction_space} illustrates the conceptual foundation of Isolation Forest applied to supply chain transactions. Normal transactions form dense clusters, while fraudulent transactions appear as isolated outliers requiring fewer partitioning steps for separation.

\subsection{Phase 2: Self-training SVM Refinement}

The second phase employs a self-training Support Vector Machine (SVM) approach to refine the fraud detection results using both labeled and unlabeled data. The SVM classifier is trained to learn the decision boundary:

\begin{equation}
f(x) = \text{sign}\left(\sum_{i=1}^{n_L} \alpha_i y_i K(x_i, x) + b\right)
\label{eq:svm_decision}
\end{equation}

The confidence measure for pseudo-labeling is based on the distance from the decision boundary:

\begin{equation}
c(x_i) = \frac{|f(x_i)|}{\max_{j} |f(x_j)|}
\label{eq:confidence}
\end{equation}

To address class imbalance during self-training, we employ a balanced pseudo-labeling strategy inspired by the work of Wei et al. \cite{wei2021crest}:

\begin{equation}
\theta_c = \theta_{base} + \beta \cdot \log\left(\frac{N_c}{N_{target}}\right)
\label{eq:class_balanced_threshold}
\end{equation}

where $N_c$ is the current number of pseudo-labeled samples for class $c$, and $N_{target}$ is the desired target number of samples for balanced representation.

We used an SVM with a radial basis function (RBF) kernel, which is well-suited for capturing nonlinear patterns in complex fraud detection tasks. The kernel parameter $\gamma$ and the regularization parameter $C$ were selected using grid search with 5-fold cross-validation performed exclusively on the labeled training subset within each fold of the main 10-fold evaluation. This nested cross-validation approach prevents data leakage while ensuring robust hyperparameter selection. The search space was defined as $C \in \{0.1, 1, 10, 100\}$ and $\gamma \in \{0.001, 0.01, 0.1, 1\}$. The combination yielding the highest average F1-score was chosen for final model training.

\begin{table*}[htb!]
\caption{Performance Comparison on DataCo Supply Chain Dataset (10-fold CV)}
\label{tab:overall_performance}
\centering
\begin{tabular}{lccccc}
\toprule
\textbf{Method} & \textbf{Precision} & \textbf{Recall} & \textbf{F1-Score} & \textbf{AUC-ROC} & \textbf{AUC-PR} \\
\midrule
Isolation Forest & 0.487 ± 0.021 & 0.524 ± 0.019 & 0.505 ± 0.018 & 0.842 ± 0.012 & 0.187 ± 0.015 \\
SVM & 0.673 ± 0.018 & 0.612 ± 0.022 & 0.641 ± 0.019 & 0.883 ± 0.010 & 0.284 ± 0.014 \\
Random Forest & 0.761 ± 0.015 & 0.694 ± 0.017 & 0.726 ± 0.014 & 0.908 ± 0.008 & 0.367 ± 0.012 \\
XGBoost & 0.782 ± 0.013 & 0.703 ± 0.016 & 0.740 ± 0.013 & 0.915 ± 0.007 & 0.389 ± 0.011 \\
Semi-supervised SVM & 0.794 ± 0.012 & 0.721 ± 0.015 & 0.756 ± 0.012 & 0.921 ± 0.006 & 0.412 ± 0.010 \\
Ensemble Method & 0.807 ± 0.011 & 0.738 ± 0.014 & 0.771 ± 0.011 & 0.926 ± 0.006 & 0.428 ± 0.009 \\
\textbf{Two-Phase (Ours)} & \textbf{0.856 ± 0.008}* & \textbf{0.782 ± 0.010}* & \textbf{0.817 ± 0.007}* & \textbf{0.943 ± 0.005}* & \textbf{0.486 ± 0.009}* \\
\bottomrule
\end{tabular}

\vspace{1mm}
{\footnotesize * Indicates statistically significant improvement over all baselines (Wilcoxon signed-rank test, $p < 0.05$).\par}
\end{table*}

\section{Experimental Setup}\label{sec4}
All experiments were conducted on Google Colab's free tier (Intel Xeon 2.20GHz CPU, 12.7GB RAM) without GPU requirements, ensuring accessibility for reproduction.

\subsection{Dataset}

We evaluated our proposed two-phase learning model on the DataCo Smart Supply Chain Dataset \cite{dataco2019}, containing 180,519 transactions (2015-2018) with 1.5\% fraud rate and 52 features covering comprehensive supply chain operations across multiple countries. We employed stratified 10-fold cross-validation with 10\% labeled data per fold, maintaining original class distribution, along with standard feature preprocessing including missing value imputation, categorical encoding, and numerical standardization.

\subsection{Evaluation Metrics}

Given the class imbalance inherent in fraud detection, we employed multiple evaluation metrics (Precision, Recall, F1-Score, AUC-ROC, AUC-PR, FPR) with statistical significance testing via Wilcoxon signed-rank test. We compared our approach against six baselines: Isolation Forest \cite{liu2008isolation}, SVM, Random Forest \cite{hernandez2024financial}, XGBoost \cite{zhou2021supply}, Semi-supervised SVM, and Ensemble Method \cite{alhashmi2023ensemble}. The Semi-supervised SVM baseline differs from our Phase 2 by operating without pre-filtering and using fixed confidence thresholds instead of our adaptive class-balanced approach (Eq.~\ref{eq:class_balanced_threshold}). All methods used identical 10\% labeled data splits, prioritizing interpretable methods suitable for real-time deployment.

\subsection{Experimental Protocol}

\subsubsection{Labeled/Unlabeled Data Split}
For our semi-supervised learning experiments, we simulated realistic scenarios with limited label availability. Within each fold of the 10-fold cross-validation, the training portion (approximately 162,467 transactions per fold) was split such that 10\% served as labeled data ($\mathcal{D}_L$ with ~16,247 samples per fold), while the remaining 90\% (~146,220 transactions per fold) formed the unlabeled set $\mathcal{D}_U$. The labeled subset maintained the original class distribution with approximately 1.5\% fraud rate. To assess the robustness of our approach under varying supervision levels, we repeated experiments with 5\%, 10\%, and 20\% labeled data ratios within each fold.

\subsubsection{Hyperparameter Settings}
The following hyperparameters were determined through systematic experimentation on a validation subset. For the Isolation Forest parameters, we set the number of trees to 100 with a subsample size of 256. The contamination factor was strategically set to 0.05 (5\%) rather than the dataset's actual fraud rate of 1.5\% to account for potential underreporting of fraud cases and to improve recall by capturing borderline anomalous transactions that may represent sophisticated fraud attempts. The sensitivity parameter $\alpha = 1.5$ (Eq.~\ref{eq:threshold}) was chosen to balance between false positives and detection accuracy. 

For self-training SVM, we set $\theta_{base} = 0.85$, $\beta = 0.3$ (Eq.~\ref{eq:class_balanced_threshold}), and $N_{target} = 0.5 \times |\mathcal{D}_L|$ per class. The algorithm runs for maximum 10 iterations with convergence when $|\mathcal{P}_t| < 50$ or F1-score change $< 0.001$.

\subsubsection{Feature Engineering Pipeline}
The 52 features from the DataCo dataset underwent comprehensive preprocessing through a four-stage pipeline.

First, missing value imputation used domain-specific strategies: median values within product categories for numerical features, mode imputation with "Unknown" category for rare categorical values, and forward-fill for sequential missing dates in temporal features.

Second, feature encoding included one-hot encoding for categorical variables with $\leq 10$ unique values, target encoding for high-cardinality features (e.g., Customer City), and cyclical encoding for temporal features (day of week, month) to preserve circularity.

Third, feature scaling normalized continuous features to zero mean and unit variance using StandardScaler, percentage-based features to 0-1 range using MinMaxScaler, and log-transformed skewed financial metrics (sales, profit) to handle outliers.

Finally, feature selection removed 3 features with $>80\%$ missing values (Customer Zipcode, Product Description, and Order Zipcode) and eliminated 2 highly correlated features (Pearson $r > 0.95$): Order Item Total and Sales per Customer (both highly correlated with Sales), resulting in 47 final features (52 - 3 - 2 = 47) that ensure data quality and computational efficiency for fraud detection.

\section{Results and Discussion}\label{sec5}

\subsection{Overall Performance Comparison}

Our two-phase approach significantly outperforms all baselines (Table \ref{tab:overall_performance}), achieving F1-score of 0.817—a 6.0\% improvement over the best baseline. Wilcoxon signed-rank tests confirm statistical significance ($p < 0.05$) across 10-fold cross-validation.

\begin{table}[htb]
\caption{Detailed Performance Analysis of Two-Phase Approach}
\label{tab:detailed_performance}
\centering
\footnotesize
\begin{tabular}{lc}
\toprule
\textbf{Metric} & \textbf{Value} \\
\midrule
Precision & 0.856 ± 0.008 \\
Recall & 0.782 ± 0.010 \\
F1-Score & 0.817 ± 0.007 \\
False Positive Rate & 0.029 ± 0.004 \\
AUC-ROC & 0.943 ± 0.005 \\
AUC-PR & 0.486 ± 0.009 \\
Training Time (seconds) & 143.5 ± 12.7 \\
Inference Time (ms/transaction) & 2.4 ± 0.4 \\
Memory Usage (GB) & 3.8 \\
Total Transactions Processed & 180,519 \\
\bottomrule
\end{tabular}
\end{table}

\subsection{Detailed Performance Analysis}

Table \ref{tab:detailed_performance} shows detailed performance metrics, demonstrating excellent precision-recall balance (0.856/0.782) with a low 2.9\% false positive rate crucial for practical deployment.

\begin{table}[htb]
\caption{Computational Efficiency Comparison}

\label{tab:efficiency_comparison}
\centering
\footnotesize
\begin{tabular}{lcccc}
\toprule
\textbf{Method} & \textbf{Train} & \textbf{Infer.} & \textbf{Mem.} & \textbf{Complex.} \\
                & \textbf{(s)} & \textbf{(ms)} & \textbf{(GB)} &  \\
\midrule
Full SVM & 432.8 & 3.8 & 11.9 & O($n^2$) \\
Semi-sup. SVM & 387.5 & 3.5 & 10.5 & O($n^2$) \\
XGBoost & 214.3 & 2.1 & 8.7 & O($n\log n$) \\
Ensemble & 298.6 & 4.2 & 10.3 & O($n\log n$) \\
\textbf{Ours} & \textbf{143.5} & \textbf{2.4} & \textbf{3.8} & \textbf{O($n\log n$)} \\
\bottomrule
\end{tabular}
\end{table}

Our approach achieves 67\% training time reduction, 68\% memory reduction (11.9GB→3.8GB), and O(n log n) scalability (Table \ref{tab:efficiency_comparison}).

\begin{table}[htb]
\caption{Performance Improvement Through Self-training Iterations on DataCo Dataset}
\label{tab:iteration_analysis}
\centering
\resizebox{\columnwidth}{!}{%
  {\tiny
  \begin{tabular}{lcccc}
  \toprule
  \textbf{Iteration} & \textbf{F1-Score} & \textbf{$\Delta$F1} & \textbf{Precision} & \textbf{Recall} \\
  \midrule
  0 (Initial) & 0.695 & - & 0.742 & 0.653 \\
  1 & 0.743 & 0.048 & 0.789 & 0.702 \\
  2 & 0.781 & 0.038 & 0.823 & 0.744 \\
  3 & 0.817 & 0.036 & 0.856 & 0.782 \\
  4 & 0.817 & 0.000 & 0.856 & 0.782 \\
  \bottomrule
  \end{tabular}
  }
}
\end{table}

\subsection{Ablation Study}

Table \ref{tab:iteration_analysis} shows the performance improvement through self-training iterations. The iterative process demonstrates consistent improvement from the initial F1-score of 0.695 to 0.817, with the largest gain (0.048) in the first iteration when high-confidence pseudo-labels are incorporated. The diminishing returns pattern (0.038, 0.036) validates our convergence criteria, achieving stability at iteration 4 without overfitting.

\subsection{Hyperparameter Sensitivity Analysis}

We evaluated the sensitivity of our approach to key hyperparameters. Table~\ref{tab:sensitivity} shows F1-score variations:

\begin{table}[htb]
\caption{Hyperparameter Sensitivity Analysis}
\label{tab:sensitivity}
\centering
\footnotesize
\begin{tabular}{lcc}
\toprule
\textbf{Parameter} & \textbf{Range Tested} & \textbf{F1-Score Range} \\
\midrule
$\alpha$ (IF threshold) & [1.0, 2.0] & 0.803 - 0.817 \\
$\theta_{base}$ (confidence) & [0.80, 0.90] & 0.809 - 0.817 \\
$\beta$ (class balance) & [0.2, 0.4] & 0.811 - 0.817 \\
\bottomrule
\end{tabular}
\end{table}

The results demonstrate robustness to parameter variations, with F1-score fluctuations within 1.7\% across reasonable parameter ranges.

\subsection{Performance with Varying Labeled Data}

As mentioned in Section 4.3, we evaluated our approach with different percentages of labeled data to assess its robustness under varying supervision levels. Table~\ref{tab:label_variation} presents the results:

\begin{table}[htb]
\caption{Performance with Varying Labeled Data Percentages}
\label{tab:label_variation}
\centering
\footnotesize
\begin{tabular}{lccc}
\toprule
\textbf{Labeled \%} & \textbf{Precision} & \textbf{Recall} & \textbf{F1-Score} \\
\midrule
5\% & 0.812 ± 0.011 & 0.738 ± 0.013 & 0.773 ± 0.010 \\
10\% & 0.856 ± 0.008 & 0.782 ± 0.010 & 0.817 ± 0.007 \\
20\% & 0.867 ± 0.007 & 0.791 ± 0.009 & 0.827 ± 0.006 \\
\bottomrule
\end{tabular}
\end{table}

Results show strong performance with only 5\% labeled data (F1-score 0.773), with diminishing returns beyond 10\% indicating effective unlabeled data utilization.

While SVM with RBF kernels provides good performance, the decision boundaries are not easily interpretable. For supply chain managers requiring explanations, we recommend extracting decision rules from the support vectors or using LIME/SHAP for post-hoc explanations. The Isolation Forest phase provides some interpretability through anomaly scores that indicate deviation from normal patterns.

\subsection{Limitations and Failure Modes}

While our approach demonstrates strong performance, several limitations warrant discussion:

\begin{itemize}
    \item Concept Drift: Our current framework assumes static fraud patterns. In practice, fraudsters continuously evolve their techniques. Future work will incorporate online learning capabilities.
    \item Failure Modes: Our approach may underperform when: (1) fraud patterns deviate significantly from anomalous behavior assumptions, (2) the unlabeled data contains a higher fraud rate than expected, affecting pseudo-labeling quality, or (3) feature distributions shift dramatically between training and deployment.
    \item Cost-Sensitive Considerations: While we report a 2.9\% false positive rate, the business impact varies by context. In high-value transactions, even this rate could be costly. Future work should incorporate domain-specific cost matrices to optimize for business objectives rather than purely statistical metrics.
\end{itemize}

\section{Conclusions}\label{sec6}

This research introduces a two-phase learning model combining Isolation Forest pre-filtering with self-training SVM refinement for supply chain fraud detection. Our approach addresses class imbalance, limited labeled data, and computational scalability challenges, achieving an F1-score of 0.817 with 2.9\% false positive rate on the DataCo dataset. The framework reduces training time by 67\% and memory usage by 73\% compared to traditional methods, demonstrating that hybrid machine learning techniques can provide robust, practical solutions for complex supply chain environments with scarce labeled data. These computational efficiency gains enable real-world deployment in large-scale systems while maintaining reliable fraud detection.

Future work will develop online learning capabilities for adapting to evolving fraud patterns and explore graph-based methods to capture complex supply chain relationships.

\section*{Data Availability}
The implementation is publicly available at: \url{https://colab.research.google.com/drive/1eIWYQbhuCgcaQ6p4JZJJ2BZhF6cQOC-z}. The DataCo dataset is accessible from Mendeley Data \cite{dataco2019}.

\end{document}